%% file: iclr2021_conference.tex
\documentclass{article} 
\usepackage{iclr2021_conference,times}

\input{math_commands.tex}

\usepackage{hyperref}
\usepackage{url}
\usepackage{graphicx}
\usepackage{amsmath}
\usepackage{mathtools}
\usepackage{textgreek}
\usepackage{amssymb}

\usepackage{color,soul}

\title{Explainable AI for Natural Adversarial Images}

\author{Tomas Folke, ZhaoBin Li$^\dagger$, Ravi B. Sojitra, Scott Cheng-Hsin Yang \& Patrick Shafto \\
Department of of Mathematics and Computer Science, Rutgers University, Newark, NJ 07102, USA \\
\texttt{\{tomas.folke, ravisoji, scott.cheng.hsin.yang, patrick.shafto\}@gmail.com} \\
$^\dagger$\texttt{liz2@carleton.edu}
}

\iclrfinalcopy 
\begin{document}

\maketitle

\begin{abstract}
Adversarial images highlight how vulnerable modern image classifiers are to perturbations outside of their training set.
Human oversight might mitigate this weakness, but depends on humans understanding the AI well enough to predict when it is likely to make a mistake.
In previous work we have found that humans tend to assume that the AI's decision process mirrors their own.
Here we evaluate if methods from explainable AI can disrupt this assumption to help participants predict AI classifications for adversarial and standard images. We find that both saliency maps and examples facilitate catching AI errors, but their effects are not additive, and saliency maps are more effective than examples. 
\end{abstract}

\section{Introduction}

\textit{Adversarial images} are images that cause the AI to be confidently wrong \citep{szegedy2013intriguing, nguyen2015deep}, despite being easily classified by humans.
Large regions of the possible input space might lead to such misclassifications \citep{goodfellow2014explaining}.
Sensitivity to adversarial images can leave an AI vulnerable to purposeful attacks \citep{eykholt2018robust}, but even in naturalistic settings some images behave "adversarially" in the sense that algorithms confidently misclassify them even though a human would not (\textit{natural adversarial images}, see \citet{hendrycks2019natural}).
It has proven challenging to build systems that are robust to adversarial images, or give low confidence to adversarial mistakes \citep{hendrycks2019natural, goodfellow2014explaining, papernot2016limitations}.
Therefore it would be helpful if humans could catch and veto such cases.
However, the default human assumption seems to be that AI classifiers share their perceptions and beliefs \citep{yang2021mitigating}. This assumption makes it harder for humans to identify adversarial cases because they themselves are not fooled by such cases \citep{papernot2016limitations, harding2018human}. Here we test whether explanations help people to predict misclassifications of natural adversarial images, which is an essential prerequisite for effective human oversight of AI systems.

A popular class of methods to explain AI is \textit{explanation-by-examples}.
Explanation-by-examples takes an AI model and its training data as inputs and selects a small subset of cases that exert high impact on the inference of the explainee.
Humans have the ability to induce principles from a few examples \citep{Mill1884,lake2020people}, which is why examples are extensively used in formal education \citep{chi1989self,aleven1997teaching,bills2006exemplification}.
The explanation-by-examples approach has many desirable properties: It is fully model-agnostic and applicable to all types of machine learning \citep{chen2018learning};
it is domain- and modality-general \citep{kanehira2019learning}; and it can be used to generate both global \citep{kim2014bayesian, vong2018} and local explanations \citep{papernot2018deep, goyal2019counterfactual}.

We have developed a computational framework for explanation-by-examples called \textit{Bayesian Teaching} \citep{yang2017explainable, vong2018}.
Based in the cognitive science of human learning \citep{Shafto2008, Shafto2014}, and drawing upon deep connections to probabilistic machine learning \citep{murphy2012machine, Eaves2016-topic}, Bayesian Teaching integrates models of human and machine learning in a single system.
Bayesian Teaching casts the problem of XAI as a problem of teaching---selecting optimal examples to teach the human user what the AI system has inferred.
The explanatory examples selected in this teaching framework have been shown to match what humans find representative of the underlying generative process \citep{Tenenbaum2001}.

We select the optimal teaching examples based on a model of the learner \citep{Shafto2014}:
\begin{align}
P_{teacher}(\sD \mid Y = c, x) \propto P_{learner}(Y = c \mid \sD, x) = \int P(Y=c \mid x, \mW)\,p(\mW \mid \sD)\,d\mW.
\label{eq:BT}
\end{align}
A quality teaching set is one that correctly helps the learner to update their prediction $Y$ of a new image $x$ belonging to the target category $c$ after learning from observed teaching set $\sD$. The target category is the category predicted by the target model, which is a ResNet-50 with pre-trained ImageNet weights.\footnote{\url{https://pytorch.org/docs/stable/torchvision/models.html}} We used Bayesian teaching to generate explanations at two levels of granularity: case-level examples and saliency maps (see Appendix for details). Then we evaluated how these two explanation features impacted human understanding of AI with adversarial examples. We evaluated human understanding by testing how well participants could predict the AI classifications for adversarial images, and compared their predictive performance relative to AI errors and correct classifications for standard images.

\section{Methods}

\subsection*{Experimental design}
The Natural Adversarial ImageNet dataset \citep{hendrycks2019natural} contains 200 categories that belong to a subset of the 1000 categories in ImageNet \citep{ILSVRC15}.
From these 200 categories, we selected 30 categories that span the spectrum of model accuracy based on ResNet-50's predictions on the standard ImageNet's validation set.
For each of these 30 categories, we made three types of trials characterized by the model's prediction on \textit{target images}: (1) model hit on an image sampled from standard ImageNet, (2) model error on an image sampled from standard ImageNet, and (3) model error on an image sampled from Natural Adversarial ImageNet.
All target images were randomly sampled from the chosen categories and dataset. We used these target images in a two alternative forced choice task, where participants were asked to predict the model classifications. The two options are referred to as the \textit{target category} and the \textit{alternative category}.

For misclassified images the target category is the model prediction, and the alternative category is the ground truth. For correctly classified images the target category is the model's predicted category, and the alternative category is the category most confusable with the target category, according to the confusion matrix constructed on ResNet-50's predictions on the standard ImageNet's validation set.
The standard images and adversarial images were matched with regards to the ground truth category of the target image, but not with regards to the target category.
For example, when the ground truth of the target image was an accordion and the AI was wrong, the target category (the AI prediction) was ``vacuum" in the standard case, but ``breastplate" in the adversarial case.

The above procedure generated 90 trials that specify the target image, the target category, and the alternative category.
These specifications were fed into the Bayesian Teaching framework, which produced a teaching set of four explanatory examples---two from the target category and two from the alternative category---for each trial.
The four examples were selected so that the learner model, once exposed to them, would infer the target image to be of the target category with probability $>0.8$.
Out of the 90 trials selected, Bayesian Teaching could not find four examples that met this criterion for one of the standard incorrect trials. Thus, the experiment had a total of 89 trials.

\subsection*{Participants}
The study protocol was approved by Rutgers University IRB.
Informed consent was obtained from all participants.
Participants were randomly allocated to four levels of explanation: (1) no explanation, (2) saliency maps only, (3) examples only, or (4) saliency maps and examples.
We tested 40 participants per condition, resulting in a total sample of 160 participants.

\subsection*{Learner model}
Because the ResNet-50 model encodes statistical patterns reflecting human labels, we adapted the ResNet-50 architecture for the learner model, which is a useful, albeit simplifying, assumption. Under the Bayesian Teaching framework, the learner model learns probabilistically. Converting the whole ResNet-50 into a probabilistic model would be computationally intractable.
Hence, we simplified the probabilistic approach by modifying only the classification layer of the learner model to make probabilistic decisions, while keeping the convolutional base deterministic. Since our aim is to teach humans a binary classification, we scaled down and converted the deterministic softmax layer, originally designed for 1000 categories, to a Bayesian logistic regression layer with two classes. 
We set a normal prior over the weights of the Bayesian classification layer for the learner model.
The normal prior is obtained by performing a Kronecker factored Laplace approximation \citep{ritter2018scalable} over the classification layer of the original ResNet-50 model, trained on ImageNet dataset over 100 epochs with data augmentation. Then, we used Laplace approximation to obtain a normal posterior over the weights of the learner, the $p(\mW \mid \sD)$ in Equation~\ref{eq:BT}. This was used in conjunction with the sigmoid likelihood $P(Y=c \mid x, \mW)$ to produce the posterior predictive $P_{learner}(Y=c \mid \sD, x)$ in Equation~\ref{eq:BT}. See Appendix A for the details.

\subsection*{Stimuli generation}
Given the specified target and alternative categories, we sampled the target images randomly from the dataset. To generate the teaching examples for each target image, we sampled $200$ teaching sets as possible candidates. Let $\{x, c\}$ be the pair of target image and label, $\sD$ be the teaching set, and $P_{learner}(Y = c \mid \sD, x)$ be the probability of the targeted prediction of the learner model. For each teaching set $\sD$, we re-initialized the prior from the two rows of the normal prior corresponding to the target and alternative categories. 
Then we trained the learner model using data augmentation over $\sD$ for $128$ epochs. Next we used Monte Carlo sampling to estimate $P_{learner}(Y = c \mid \sD, x)$. The first teaching set for which $P_{learner}(Y = c \mid \sD, x)>0.8$ was selected as a quality teaching set for the experiment. 
The saliency map for each image (including both target and explanatory examples) is generated following the same procedure as described in \citep{yang2021mitigating} (see Appendix B).

\section{Results}
\begin{figure}[ht!]
  \centering
  \includegraphics[width=\columnwidth]{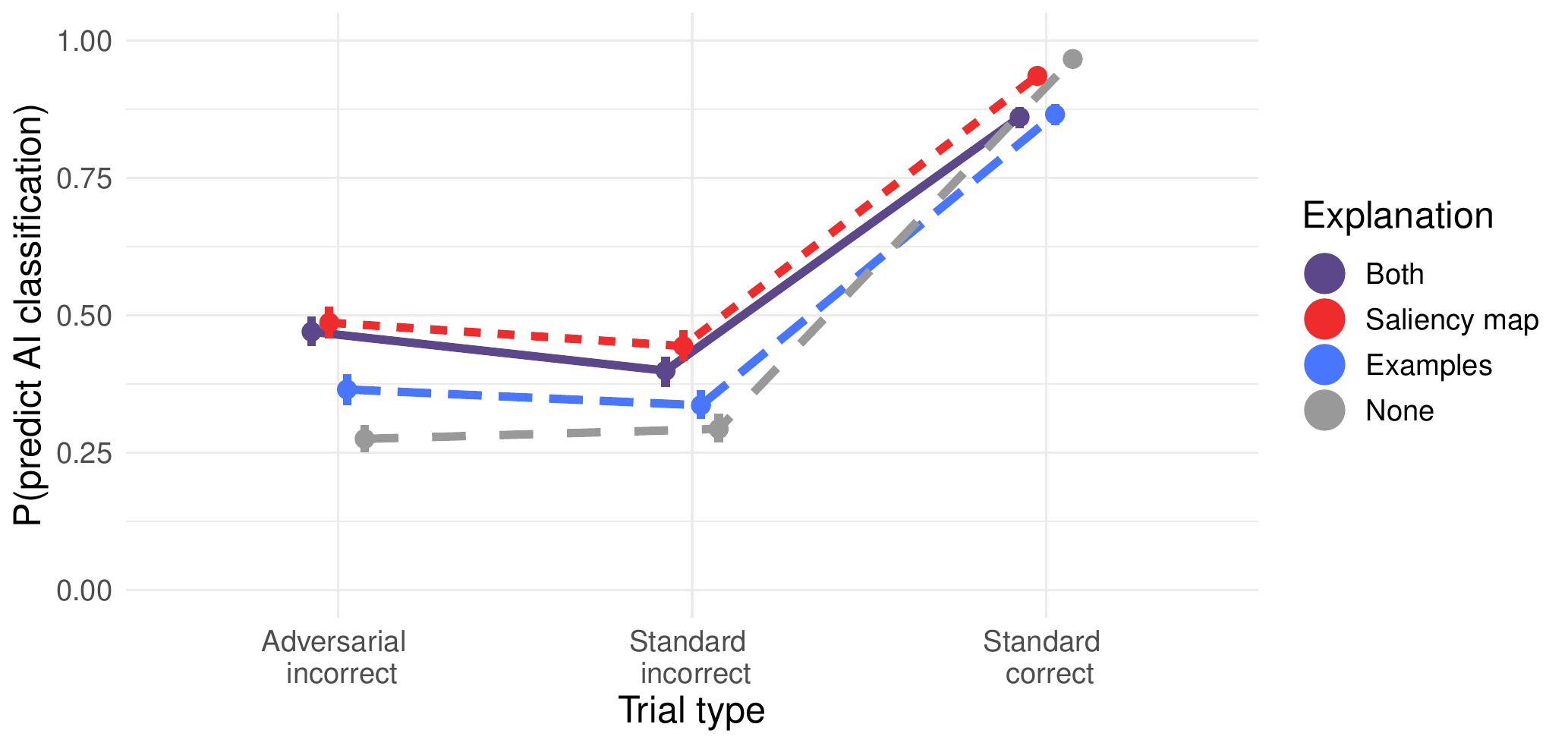}
  \caption{Explanations improve predictive performance for AI mistakes. Saliency maps are more beneficial than examples. The effects of the two explanation types are not additive, as combined explanations are associated with slightly worse performance than the saliency maps alone. Error bars signify 95\% bootstrapped confidence intervals.} 
  \label{fig:explanation_effectiveness}
\end{figure}

We aim to determine how well humans can predict AI classifications as a function of whether the target image is adversarial, and what explanation features they have access to.
To test this we first compared the performance of three nested logistic hierarchical regressions.
The simplest model represents the null hypothesis that predictive accuracy differed between participants, target categories, and trial types, but that the explanations did not impact predictive performance.
This \textit{null model} was formalized such that human predictive accuracy at the trial level was based on two random intercepts based on participant and target category, respectively, and a fixed effect of trial type (standard correct, standard incorrect, and adversarial incorrect; treating adversarial incorrect as the reference condition). 
The second model represents the hypothesis that the explanations impacted the predictive performance of the participants, but that explanation effectiveness was constant across trial types.
This \textit{explanation model} expanded on the null model by adding main effects for whether participants were exposed to saliency map explanations and example explanations.
The final model represented the hypothesis that the impact of the two explanation features (examples and saliency maps) were not additive, and that they varied between trial types.
This \textit{interaction model} built on the explanation model by adding interaction terms for the two explanation features and the trial types.
The explanation model captured prediction accuracy better than the null model according to a likelihood ratio test ($\chi^2$(2) = 45.37, p $<$ .0001), and the interaction model outperformed the explanation model ($\chi^2$(7) = 209.62, p $<$ .0001). These results are consistent with the hypothesis that explanations did impact performance differently for different trial types.
To explore these effects more thoroughly we studied the coefficients of the interaction model, see Table~\ref{table:coefficients}, and Figure~\ref{fig:explanation_effectiveness}.

\begin{table}
\begin{center}
\caption{Coefficients of the interaction model}
\label{table:coefficients}
\end{center}
\begin{tabular}{l l l}
\hline
 & Coefficient 
 & SE \\
\hline
1. (Intercept)                                                     & $-1.06^{***}$ 
& $(0.14)$ \\
2. Trial type: Standard incorrect                              
& $0.09$
& $(0.10)$ \\
3. Trial type: Standard correct
& $4.64^{***}$
& $(0.18)$ \\
4. Saliency maps present
& $1.01^{***}$
& $(0.12)$ \\
5. Examples present
& $0.46^{***}$
& $(0.12)$ \\
6. Trial type: Standard incorrect $\times$ Saliency maps present
& $-0.29^{*}$
& $(0.13)$ \\
7. Trial type: Standard correct $\times$ Saliency maps present
& $-1.71^{***}$
& $(0.22)$ \\
8. Trial type: Standard incorrect $\times$ Examples present
& $-0.24$
& $(0.13)$ \\
9. Trial type: Standard correct $\times$ Examples present
& $-2.01^{***}$
& $(0.21)$ \\
10. Saliency maps present $\times$ Examples present
& $-0.53^{**}$
& $(0.17)$ \\
11. Trial type: Standard incorrect $\times$ Saliency maps present $\times$ Examples present
& $0.11$
& $(0.18)$ \\
12. Trial type: Standard correct $\times$ Saliency maps present $\times$ Examples present
& $1.18^{***}$
& $(0.27)$ \\
\hline
\multicolumn{2}{l}{\scriptsize{$^{***}p<0.001$; $^{**}p<0.01$; $^{*}p<0.05$}}
\end{tabular}
\end{table}

Absent intervention, human predictive accuracy is similar for standard incorrect trials and adversarial incorrect trials, but much higher for standard correct images.
This may imply that absent explanations, humans tend to assume that the AI makes correct classifications, in line with previous work showing that the default human assumption is that AI classifications will match their own \citep{yang2021mitigating}.
While both saliency maps and examples significantly improve predictive performance on adversarial images, this improvement is about four times larger for the saliency maps. Additionally, the effect of the two explanation features are not additive.
Comparing the relative benefit of explanations on standard incorrect trials versus adversarial trials, we note that the improvement from saliency maps is significantly smaller for standard trials relative to adversarial trials.
The improvement from examples is also smaller for standard incorrect trials, but not significantly so.  
Finally, for the standard correct trials all interventions are associated with a decrease in performance.

\section{Discussion}
In this paper we tested whether explanations generated by Bayesian teaching help humans predict AI classifications for standard and adversarial images.
We found that explanations, and saliency maps in particular, improved participants' predictive accuracy for AI mistakes.
We also learned that saliency maps were better at alerting participants to adversarial (as opposed to standard) misclassifications, presumably because saliency maps show when the classifier attends to strange features, as it tends to do in adversarial cases.
In this study we focused on natural adversarial cases for two reasons: 1) We expect they pose a harder prediction problem for humans, relative to artificial adversarial images that are often distorted in ways that humans can detect, 2) We expect that natural adversarial images help users become aware of failure modes of the AI model outside of the training set. As such, adversarial images, together with explanations, may help users develop more sophisticated mental models of an AI's decision rules. 


\input{iclr2021_conference.bbl}
\bibliographystyle{iclr2021_conference}

\appendix
\section{Appendix}

To give an overview, Section~\ref{sec:posterior_predictive} describes the computation of the posterior predictive given the softmax likelihood and the posterior on the weights of the classification layer; see Equation~\ref{eq:posterior_predictive}. Section \ref{sec:approximate_posterior} describes the approximation used to obtain the posterior on the weights, which is obtained from the softmax likelihood and a normal prior on the weights; see Equation~\ref{eq:posterior}. Section~\ref{sec:normal_prior} describes the construction of the normal prior, which uses the weights obtained by training the classification layer of the RestNet-50 on ImageNet as the mean, and the hessian of the log-likelihood loss function as the precision matrix. These components complete the specifications of the learner model.

\subsection{Posterior Predictive}\label{sec:posterior_predictive}
Following \citep{murphy2012machine}, we used Monte Carlo integration to estimate the learner model's posterior predictive, $P(Y=c \mid x, \sD)$. First, we trained the model on a dataset $\sD = \{\evd_1,\dots, \evd_n\} = \{(\evx_1, \evy_1),\dots, (\evx_n, \evy_n)\}$ to learn the weights $\mW$ for the classification layer while retaining the pre-trained weights for the convolutional base. Given a new datapoint $x$ and label $c$, the predictive probability on $Y = c$ is:

\begin{align}
P(Y=c \mid x, \sD) &= \int P(Y=c \mid x,\sD,\mW)\,p(\mW \mid x, \sD)\,d\mW \nonumber\\
&=\int P(Y=c \mid x,\mW)\,p(\mW \mid \sD)\,d\mW \label{eq:posterior_predictive}\\
&\approx \frac{1}{s} \sum_{m=1}^s P(Y=c \mid x,\mW_{m}), \nonumber
\end{align}
where $\mW_{m}$ are samples from the normal posterior obtained using Laplace approximation. For both models, we set $s = 100$, i.e.~ we sampled a set of $100$ weights per image.

\subsubsection*{Model implementation}
For multinomial logistic regression on the learner model, $\vw_c$ represents the weights for class $c$ in all classes $\sC$, and the predictive probability is:

$$P(Y=c \mid x, \sD) \approx \frac{1}{s} \sum_{m=1}^s \frac{\exp(\vw_{m, c}^T x)}{\sum_{c' \in \sC}\exp(\vw_{m, c'}^T x)}$$

\subsection{Approximating the posterior}\label{sec:approximate_posterior}
Following \citep{murphy2012machine}, we used Laplace approximation to obtain the posterior of the weights $p(\mW \mid \sD)$ for the classification layer. The posterior for a model trained on dataset $\sD$ is given by:

\begin{align}
p(\mW \mid \sD) &= \frac{P(\sD \mid \mW)\,p(\mW)}{P(\sD)} \label{eq:posterior}\\
&= \frac{1}{P(\sD)}e^{\log(P(\sD \mid \mW)\,p(\mW))} \nonumber.
\end{align}

To obtain the mode of the posterior, we define a loss function $\newterm{L(\mW)} \triangleq -\log(P(\sD \mid \mW)p(\mW))$ and solve for:

\begin{align*}
\mW^* = \argmin_\mW L(\mW).
\end{align*}

The mode $\mW^*$ will be used as the mean of the approximate posterior, which is set to be a normal distribution. To obtain the covariance matrix for the posterior normal, we perform a second-order Taylor expansion around $\mW^*$ whereby:

\begin{align*}
L(\mW) \approx L(\mW^*) - (\mW-\mW^*)\left.\frac{\partial L(\mW)}{\partial \mW}\right|_{\mW=\mW^*} + \frac{1}{2} (\mW-\mW^*)^T \left.\frac{\partial^2 L(\mW)}{\partial \mW^2}\right|_{\mW=\mW^*} (\mW-\mW^*).
\end{align*}

Because $\mW^*$ is the mode (i.e.,~ the maxima), the gradient $\frac{\partial L(\mW)}{\partial \mW}|_{\mW=\mW^*}$ is zero. Define the hessian of the loss $\mH \triangleq \frac{\partial^2 L(\mW)}{\partial \mW^2}|_{\mW=\mW^*}$. The posterior then becomes:

\begin{align*}
p(\mW \mid \sD) &= \frac{1}{P(\sD)}\exp[-L(\mW^*) - \frac{1}{2} (\mW-\mW^*)^T \mH (\mW-\mW^*)] \\
&= \frac{e^{-L(\mW^*)}}{P(\sD)}\exp[- \frac{1}{2} (\mW-\mW^*)^T \mH (\mW-\mW^*)].
\end{align*}

By presuming $P(\sD) = e^{-L(\mW^*)} * \sqrt{(2 \pi)^k |\mH^{-1}|}$, where $k$ is the size of the input for the classification layer, (i.e.,~ the length of the feature vector after getting transformed by the convolutional base), $p(\mW \mid \sD)$ is equivalent to a multivariate normal distribution whereby:

\begin{align}
\mW \sim \mathcal{N}(\mW^*,\mH^{-1}).
\nonumber 
\end{align}

\subsubsection*{Model implementation}

The negative log likelihood is given by:

\begin{align*}
-\log{P(\sD \mid \mW)} &= -\log{\prod_i \prod_c P(Y_i=c|x_i, \mW)^{\1_\mathrm{Y_i = c}}} \\
&= -\sum_i \sum_c  \1_\mathrm{Y_i = c} \log\left[\frac{\exp(\vw_c^T x_i)}{\sum_{c' \in C}\exp(\vw_{c'}^T x_i)}\right] \\
&= -\sum_i \left[\sum_c \1_\mathrm{Y_i = c} \vw_{c}^T x_i - \log\left(\sum_{c' \in \sC}\exp(\vw_{c'}^T x_i)\right)\right].
\end{align*}

Therefore, given prior $\mW_0 \sim \mathcal{N} (\mM_0,\mSigma_0)$, $L(\mW)$ is:

\begin{align}
L(\mW) &= -\log(P(\sD \mid \mW)\,p(\mW_0)) \nonumber\\
&= -\log{P(\sD\mid \mW)} - \log{p(\mW_0)} \nonumber\\
&= \sum_i \left[\sum_c \1_\mathrm{y_i = c} \vw_{c}^T x_i - \log(\sum_{c' \in C}\exp{\vw_{c'}^T x_i})\right] \nonumber\\
& +\log(\sqrt{(2 \pi)^k |\mSigma_0|}) + (\mW-\mM_0)^T \mSigma_0^{-1} (\mW-\mM_0). \label{eq:loss}
\end{align}

For the learner model, we solved for $\mW^* = \argmin_\mW L(\mW)$ using PyTorch L-BFGS optimizer because the size of the training set is small for the learner model. Also, we computed the hessian $\mH$ using a hessian solver built upon PyTorch.\footnote{\url{https://github.com/mariogeiger/hessian}} This learner model is then used to obtain the teaching set $\sD$ as explanatory examples as described in the main text.

\subsection{Normal prior on learner model}\label{sec:normal_prior}

We used a normal distribution for the prior on the weights for the learner model: $\mathcal{N}(\mW_0,\Sigma_0=\mH_0^{-1})$. For the mean of the prior $\mW_0$, we used the weights obtained from training the classification layer of ResNet-50 on ImageNet over 100 epochs with data augmentation.\footnote{\url{https://github.com/pytorch/examples/blob/d587b53f3604b029764f8c864b6831d0ab269008/imagenet/main.py}}
For the precision matrix $\Sigma_0^{-1}=\mH_0$, we aimed to use the hessian of a loss function that has the same form as Equation~\ref{eq:loss}, but with different values for the likelihood and prior. However, we did not evaluate the hessian directly---the classification layer of ResNet-50 has 1024 input features and 1000 classes, resulting in a $(1024\times 1000) \times (1024\times 1000)$ dimensional $\mH_0$ matrix, which cannot be computed or stored. Therefore, we used Kronecker factored Laplace approximation to estimate $\mH_0$, as detailed in the next section, following the work in \citep{ritter2018scalable}. 

For the rest of Appendix A, we will simplify the notation to $\mW_0 \rightarrow \mW$ and $\mH_0 \rightarrow \mH$. Thus, in the following subsections $\mW$ now refers to the learner model's prior weights as opposed to the posterior weights, and $\mH$ refers to the learner model's precision matrix on the normal prior as opposed to that on the normal posterior. Also, $\sD$ now refers to the entire ImageNet dataset as opposed to the teaching set.

\subsubsection*{Kronecker Factored Laplace approximation}

For the classification layer of ResNet-50, let the input feature vector (including the bias term) be $\vz = \{\evz_i, \dots, \evz_m\}$ and the pre-activation vector be $\va = \{\eva_i, \dots, \eva_n\}$ (i.e., the class activations before applying the softmax function). The $m \times n$ weight matrix $\mW$, whereby the $i$th row of $\mW$ is the weight vector $\vw_i$ for class $i$, connects the two layers with $\va = \mW \vz$. 

Note that we can write the hessian of the loss $\newterm{\mH} \triangleq \frac{\partial^2 L(\mW)}{\partial \mW^2}|_{\mW=\mW^*}$ as a sum of individual hessian of the loss per data point $\sD_i$ (without the prior) and the negative log prior:

\begin{align*}
\mH &= \frac{\partial^2 L(\mW)}{\partial \mW^2} \\
&= - \frac{\partial^2}{\partial \mW^2} \log (P(\sD\mid \mW)\,p(\mW)) \\
&= - \frac{\partial^2}{\partial \mW^2} [ \log P(\sD\mid \mW) + \log p(\mW) ] \\
&= - \frac{\partial^2}{\partial \mW^2} \left[ \log \prod_i P(\sD_i\mid \mW) + \log p(\mW) \right] \\
&= - \frac{\partial^2 }{\partial \mW^2} \left[ \sum_i \log P(\sD_i \mid \mW) + \log p(\mW) \right] \\
&= \sum_i \frac{\partial^2 L_i(\mW)}{\partial \mW^2} - \frac{\partial^2 \log p(\mW)}{\partial \mW^2}.
\end{align*}

The first derivative of the individual loss per data point $L_i(\mW)$, with respect to a weight $\emW_{i, j}$ connecting an input feature $\evz_j$ and a pre-activation node $\eva_i$ is:

\begin{align*}
\frac{\partial L_i(\mW)}{\partial \emW_{i, j}}  &= \frac{\partial L_i(\mW)}{\partial \eva_i} \frac{\partial \eva_i}{\partial \emW_{i, j}} \\ 
&= \evz_j \frac{\partial L_i(\mW)}{\partial \eva_i}.
\end{align*}

The second derivative with respect to another weight $\emW_{k, l}$ is:

\begin{align*}
\frac{\partial}{\partial \emW_{k, l}} \frac{\partial L_i(\mW)}{\partial \emW_{i, j}}   &= \frac{\partial}{\partial \emW_{k, l}}  
\evz_j \frac{\partial L_i(\mW)}{\partial \eva_i} \\
&= \evz_j \frac{\partial}{\partial \eva_i} \frac{\partial L_i(\mW)}{\partial \emW_{k, l}} \\
&= \evz_j \evz_l \frac{\partial^2 L_i(\mW)}{\partial \eva_i \partial \eva_k}.
\end{align*}

We can express the hessian of $L_i(\mW)$ over $\mW$ using a Kronecker product. Let $\newterm{\vw} \triangleq vec(\mW)$, where the $vec$ operator stacks the columns of $\mW$ into a vector. Also, define $\newterm{\mZ} \triangleq \vz\vz^T$ to be the outer product of $\vz$, and $\newterm{\mA}$ to be the hessian of $L_i(\mW)$ over $\va$ such that $\mA_{i,j} \triangleq \frac{\partial^2 L_i(\mW)}{\partial \eva_i \partial \eva_j}$. Then: 

$$\frac{\partial^2 L_i(\mW)}{\partial \vw^2} = \mZ_i \otimes \mA_i\;.$$

We set the prior precision to be $\tau \mI$, where $\tau$ is a constant controlling the precision.\footnote{This corresponds to L2 regularization. Also, this is the prior for training the classification layer of the original ResNet-50, not the prior on the learner model's weights.} Then the negative log prior is:

\begin{align*}
- \frac{\partial^2}{\partial \vw^2} \log P(\vw) &= \frac{\partial^2}{\partial \vw^2} \left[\log (\sqrt{(2 \pi)^k |\tau^{-1}\mI|}) + \frac{1}{2}(\vw-\vw_0)^T \tau\mI (\vw-\vw_0)\right] \\
&= \tau \mI \frac{\partial^2}{\partial \vw^2} \left[\frac{1}{2} || \vw-\vw_0 || \right] \\
&= \tau \mI \frac{\partial}{\partial \vw} (\vw-\vw_0) \\
&= \tau \mI. 
\end{align*}

Now we could express $\mH$ as a Kronecker product. Using probability notation to denote $\E[\mZ]$ and $\E[\mA]$ as the mean of $\mZ_i$ and $\mA_i$ over the whole dataset of size $n$, and presuming that $\vz$ and $\va$ are independent, $\mH$ is:

\begin{align*}
\mH &= \sum_i \frac{\partial^2 L_i(\mW)}{\partial \vw^2} - \frac{\partial^2 \log P(\vw)}{\partial \vw^2} \\
&= \sum_i \mZ_i \otimes \mA_i + \tau \mI \\
&= n \E[\mZ_i \otimes \mA_i] + \tau \mI \\
&= n [\E[\mZ_i] \otimes \E[\mA_i]] + \tau \mI \\
&= \left(\sqrt{n} \E[\mZ_i]\right) \otimes \left(\sqrt{n} \E[\mA_i]\right) + \tau \mI.
\end{align*}

To incorporate $\tau \mI$ into the Kronecker product: 

\begin{align*}
\mH &= \left(\sqrt{n} \E[\mZ_i]\right) \otimes \left(\sqrt{n} \E[\mA_i]\right) + \tau \mI \\
&\approx \left(\sqrt{n} \E[\mZ_i] + \sqrt{\tau} \mI\right) \otimes \left(\sqrt{n} \E[\mA_i] + \sqrt{\tau} \mI\right).
\end{align*}

Considering that $n$ is large ($1$ million for Imagenet) and that the prior is weak, the regularization effect is negligible. Hence, we set $\tau$ to $0$.

\subsubsection*{Matrix Normal Posterior}
Once we have expressed $\mH$ as a Kronecker product, we can express the learner model's prior of the weights (which confusingly is a posterior itself from training on the ImageNet dataset) as a matrix normal distribution, where the covariance is broken down into two manageable $1024\times 1024$ and $1000\times 1000$ dimension matrices.

Considering that the inverse of a Kronecker product is the Kronecker product of the inverses and defining $\newterm{\mU} \triangleq \sqrt{n} \E[\mZ_i] + \sqrt{\tau} \mI$ and $\newterm{\mV} \triangleq \sqrt{n} \E[\mA_i] + \sqrt{\tau} \mI$:

$$\mH^{-1} = (\mU \otimes \mV)^{-1} = \mU^{-1} \otimes \mV^{-1}$$

Then\footnote{We believed that in the original paper \citep{ritter2018scalable} the authors mistakenly swapped $\mU$ and $\mV$.}: 

$$\vw \sim \mathcal{N}(\vw^*,\mH^{-1}) \iff \vw \sim \mathcal{N}(\vw^*,\mU^{-1} \otimes \mV^{-1}) \iff \mW \sim \mathcal{MN}(\mW^*,\mU^{-1}, \mV^{-1})$$ 

We can sample from the matrix normal distribution using Cholesky decomposition. Letting $\vu$ be the lower triangular matrix of $\mU^{-1} = \vu\vu^T$, $\vv$ be the upper triangular matrix of $\mV^{-1} = \vv^T\vv$, and $\rmQ \sim \mathcal{MN}(0,\,\mI,\,\mI)$ be the standard normal distribution (which can be sampled from a univariate standard normal distribution and reshaped into an $m\times n$ matrix):

$$\mW_s = \mW^* + \vu\rmQ_s\vv$$

The validation accuracy of the Kronecker factored probabilistic ResNet-50 on ImageNet, using Monte Carlo sampling drawn from the matrix normal posterior, is 75.9\% for top-1 and 92.8\% for top-5, close to the original ResNet-50 accuracy.

\section{Appendix}
Following \citep{yang2021mitigating}, we generated saliency maps by using Bayesian Teaching to select \textit{pixels} of an image that help a learner model to the targeted prediction. Let $q_{teacher}(m \mid Y=c, x)$ be the probability that a mask $m$ will lead the learner model to predict the image $x$ to be in category $c$ when the mask is applied to the image. This is expressed by Bayes' rule as

\begin{align*}
  q_{teacher}(m \mid Y=c, x) = \frac{Q_{learner}(Y=c \mid x, m) p(m)}
  {\int_{\Omega_M} Q_{learner}(Y=c \mid x, m) p(m)}.
\end{align*}

Here, $Q_{learner}(Y=c \mid x, m)$ is the probability that the ResNet-50 model with pre-trained ImageNet weights will predict the $x$ masked by $m$ to be $c$; $p(m)$ is the prior probability of $m$; and $\Omega_M = [0, 1]^{W \times H}$ is the space of all possible masks on an image with $W\times H$ pixels. We used a sigmoid-function squashed Gaussian process prior for $p(m)$.

Instead of sampling the saliency maps directly from the above equation, we find the expected saliency map for each image by Monte Carlo integration:

\begin{align}
\text{E}[M \mid x,c]
&=\int_{\Omega_M} m\ q_{teacher}(m \mid Y=c, x) \nonumber\\
&\approx \frac{\sum_{i=1}^N m_i\ Q_{learner}(Y=c \mid x, m_i)}{\sum_{i=1}^N Q_{learner}(Y=c \mid x, m_i)},
\label{eq:map}
\end{align}
where $m_i$ are samples from the prior distribution $p(m)$, and $N=1000$ is the number of Monte Carlo samples used. The expected mask is used as the saliency map.

\subsection{Implementation}
To generate the saliency map for an image $x$, we first resized $x$ to be 224-by-224 pixels. A set of 1000 2D functions were sampled from a 2D Gaussian process (GP) with an overall variance of $100$, a constant mean of $-100$, and a radial-basis-function kernel with length scale 22.4 pixels in both dimensions. The sampled functions were evaluated on a 224-by-224 grid, and the function values were mostly in the range of $[-500,300]$. A sigmoid function, $1 / (1 + \exp(-a))$, was applied to the sampled functions to transform each of the function values $a$ to be within the range $[0,1]$. This resulted in 1000 masks. The mean of the GP controlled how many effective zeros there were in the mask, and the variance of the GP determined how fast neighboring pixel values in the mask changed from zero to one. The 1000 masks were the $m_i$'s in Equation~\ref{eq:map}. We produced 1000 masked images by element-wise multiplying the image $x$ with each of the masks. The term $Q_{learner}(Y=c \mid x, m_i)$ was the ResNet-50's predictive probability that the $i^\textrm{th}$ masked image was in category $c$. Having obtained these predictive probabilities, we averaged the 1000 masks according to Equation~\ref{eq:map} to produce the saliency map of image $x$.

\end{document}

%% file: math_commands.tex
\usepackage{amsmath,amsfonts,bm}

\newcommand{\newterm}[1]{{\bf #1}}

\def\eqref#1{equation~\ref{#1}}

\def\1{\bm{1}}

\def\rmQ{{\mathbf{Q}}}

\def\va{{\bm{a}}}

\def\vu{{\bm{u}}}
\def\vv{{\bm{v}}}
\def\vw{{\bm{w}}}

\def\vz{{\bm{z}}}

\def\eva{{a}}

\def\evd{{d}}

\def\evx{{x}}
\def\evy{{y}}
\def\evz{{z}}

\def\mA{{\bm{A}}}

\def\mH{{\bm{H}}}
\def\mI{{\bm{I}}}

\def\mM{{\bm{M}}}

\def\mU{{\bm{U}}}
\def\mV{{\bm{V}}}
\def\mW{{\bm{W}}}

\def\mZ{{\bm{Z}}}

\def\mSigma{{\bm{\Sigma}}}

\DeclareMathAlphabet{\mathsfit}{\encodingdefault}{\sfdefault}{m}{sl}
\SetMathAlphabet{\mathsfit}{bold}{\encodingdefault}{\sfdefault}{bx}{n}

\def\sC{{\mathbb{C}}}
\def\sD{{\mathbb{D}}}

\def\emW{{W}}

\newcommand{\E}{\mathbb{E}}

\DeclareMathOperator*{\argmin}{arg\,min}